# Automated Coin Recognition System using ANN

Shatrughan Modi
Dept. of Computer Science and Engineering
Thapar University
Patiala-147004, India.

Dr. Seema Bawa
Dept. of Computer Science and Engineering
Thapar University
Patiala-147004, India.

## ABSTRACT
Coins are integral part of our day to day life. We use coins everywhere like grocery store, banks, buses, trains etc. So it becomes a basic need that coins can be sorted and counted automatically. For this it is necessary that coins can be recognized automatically. In this paper we have developed an ANN (Artificial Neural Network) based Automated Coin Recognition System for the recognition of Indian Coins of denomination ₹1, ₹2, ₹5 and ₹10 with rotation invariance. We have taken images from both sides of coin. So this system is capable of recognizing coins from both sides. Features are extracted from images using techniques of Hough Transformation, Pattern Averaging etc. Then, the extracted features are passed as input to a trained Neural Network. 97.74% recognition rate has been achieved during the experiments i.e. only 2.26% miss recognition, which is quite encouraging.

## General Terms
Neural Network, Pattern Recognition, Image Processing.

## Keywords
Pattern Averaging, Hough Transform for circle detection, Automated Coin Recognition.

## 1. INTRODUCTION
We can not imagine our life without coins. We use coins in our daily life almost everywhere like in banks, supermarkets, grocery stores etc. They have been the integral part of our day to day life. So there is basic need of highly accurate and efficient automatic coin recognition system. In-spite of daily uses coin recognition systems can also be used for the research purpose by the institutes or organizations that deal with the ancient coins. There are three types of coin recognition systems available in the market based on different methods:

- Mechanical method based systems
- Electromagnetic method based systems
- Image processing based systems

The mechanical method based systems use parameters like diameter or radius, thickness, weight and magnetism of the coin to differentiate between the coins. But these parameters can not be used to differentiate between the different materials of the coins. It means if we provide two coins one original and other fake having same diameter, thickness, weight and magnetism but with different materials to mechanical method based coin recognition system then it will treat both the coins as original coin so these systems can be fooled easily.

The electromagnetic method based systems can differentiate between different materials because in these systems the coins are passed through an oscillating coil at a certain frequency and different materials bring different changes in the amplitude and direction of frequency. So these changes and the other parameters like diameter, thickness, weight and magnetism can be used to differentiate between coins. The electromagnetic method based coin recognition systems improve the accuracy of recognition but still they can be fooled by some game coins.

In the recent years coin recognition systems based on images have also come into picture. In these systems first of all the image of the coin to be recognized is taken either by camera or by some scanning. Then these images are processed by using various techniques of image processing like FFT [1, 2], Gabor Wavelets [3], DCT, edge detection, segmentation, image subtraction [4], decision trees [5] etc and various features are extracted from the images. Then based on these features different coins are recognized.

## 2. RELATED WORK
In 1992 [6] *Minoru Fukumi et al.* presented a rotational invariant neural pattern recognition system for coin recognition. They performed experiments using 500 yen coin and 500 won coin. In this work they have created a multilayered neural network and a preprocessor consisting of many slabs of neurons to provide rotation invariance. They further extended their work in 1993 [7] and tried to achieve 100% accuracy for coins. In this work they have used BP (Back Propagation) and GA (Genetic Algorithm) to design neural network for coin recognition. *Adnan Khashman et al.* [8] presented an Intelligent Coin Identification System (ICIS) in 2006. ICIS uses neural network and pattern averaging for recognizing rotated coins at various degrees. It shows 96.3% correct identification i.e. 77 out of 80 variably rotated coin images were correctly identified. Mohamed Roushdy [9] had used Generalized Hough Transform to detect coins in image.

In our work we have combined Hough Transform and Pattern Averaging to extract features from image. Then, these features are used to recognize the coins. In section 3 implementation details are given. In section 4 we have presented training and testing data. Then, in section 5 the experimental results are provided. Then, in section 6 we have concluded the work.

## 3. IMPLEMENTATION DETAILS
Coin recognition process has been divided into seven steps. The architecture of Automated Coin Recognition System is shown in Fig. 1.

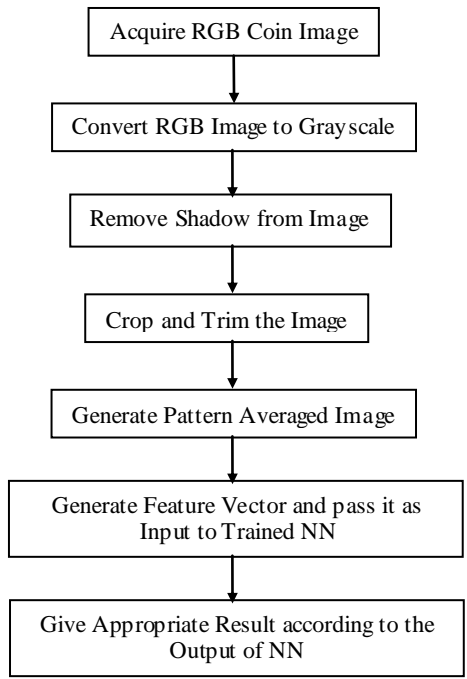

**Fig. 1: Architecture for Automated Coin Recognition System**

## 3.1 Acquire RGB Coin Image
This is the first step of coin recognition process. In this step the RGB coin image is acquired. Indian coins of denominations ₹1, ₹2, ₹5 and ₹10 were scanned from both sides at 300 dpi (dots per inch) using color scanner as shown in Fig. 2. Five coins of each denomination were scanned.

## 3.2 Convert RGB Coin Image to Grayscale
From the first step the image we got is a 24-bit RGB image. Image processing of colored images takes more time than the grayscale images. So, to reduce the time required for processing of images in further steps it is good to convert the 24-bit RGB image to 8-bit Grayscale image.

## 3.3 Remove Shadow of Coin from Image
In this step, shadow of the coin from the Grayscale image is removed. As all the coins have circular boundary. So, for removing shadow Hough Transform for Circle Detection [9] is used. For this first of all edge of the coin is detected using Sobel Edge Detection. Following is the pseudo code for Hough Transform:

Step 1. Define a 3-dimensional Hough Matrix of (M × N × R), where M, N is the height and width of the Grayscale image and R is the no. of radii for which we want to search.
Step 2. For each edge pixel (x, y) and for particular radius r, search circle center coordinates (u, v) that satisfy the equation $(x-u)^2+(y-v)^2=r^2$ and increase count in Hough Matrix at (u, v, r) by 1.
Step 3. Repeat step 2 for other radii.
Step 4. Find the maximum value from the Hough Matrix. The corresponding indices give the center coordinates and radius of coin.

Now based on the center coordinates and radius, the coin is extracted from the background. So, in this way the shadow of the coin is removed. Fig. 3 shows a coin with shadow and Fig. 4 shows the coin without shadow after applying Hough Transform.

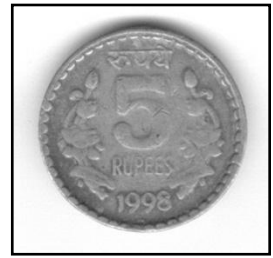 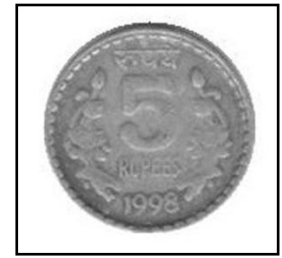

**Fig. 3: Image with Shadow**   **Fig. 4: Image without Shadow**

## 3.4 Crop and Trim the Image
After shadow removal the image is cropped so that we just have the coin in the image. Then after cropping, coin image is trimmed to make it of equal dimension of 100 × 100.

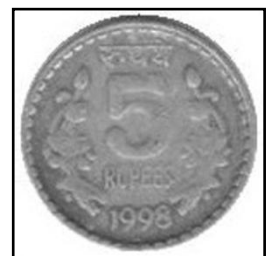 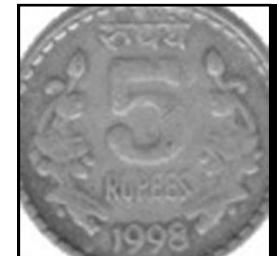

**Fig. 5: Cropped Image**       **Fig. 6: 100×100 Trimmed Image**

## 3.5 Generate Pattern Averaged Image
The 100×100 trimmed coin images become the input for the trained neural network. But to reduce the computation and complexity in the neural network these images are further reduced to size 20×20 by segmenting the image using segments of size 5×5 pixels, and then taking the average of pixel values within the segment. This can be represented by mathematical equations, as shown in (1) and (2):

$$Sum_i = \sum_{j=1}^{5}\sum_{k=1}^{5} P_{ijk} \qquad \ldots(1)$$

$$SegAvg_i = Sum_i / 25 \qquad \ldots(2)$$

where *i, j, k* is the segment no., row no. and column no. of a particular segment respectively, $Sum_i$ is the sum of the pixel values $P_{ijk}$ of the segment *i*, $SegAvg_i$ is the average of pixel values of segment *i*.

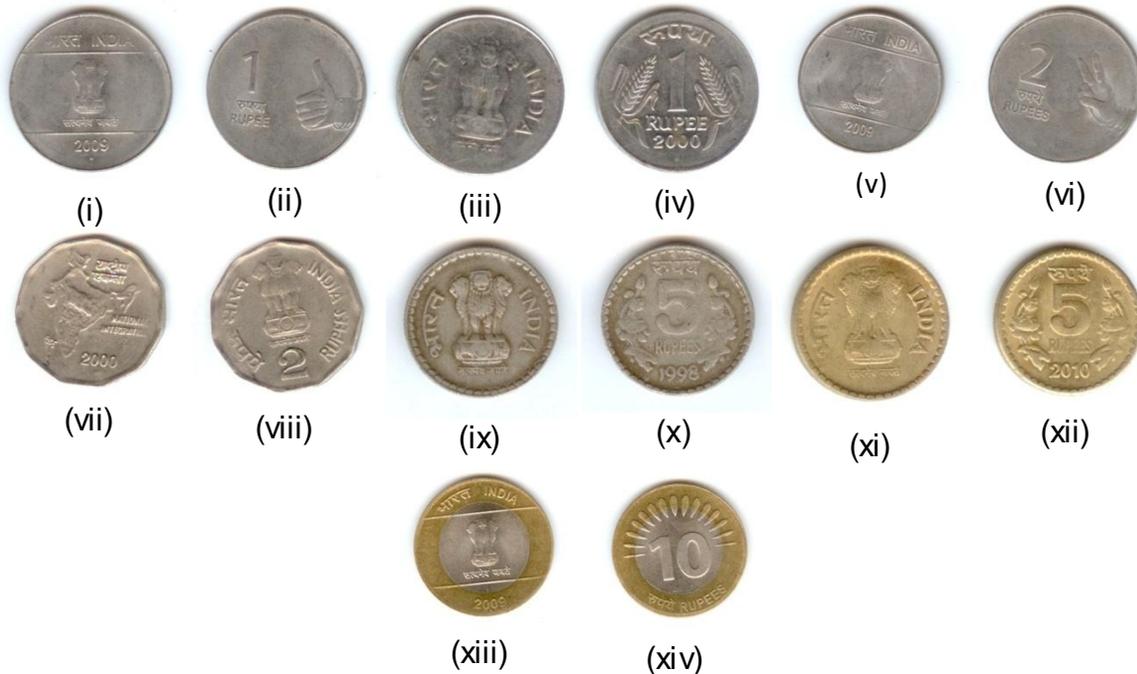

**Fig. 2: Indian Coins of different Denominations; (i) Head of ₹1 coin (1st type), (ii) Tail of ₹1 coin (1st type), (iii) Head of ₹1 coin (2nd type), (iv) Tail of ₹1 coin (2nd type), (v) Head of ₹2 coin (1st type), (vi) Tail of ₹2 coin (1st type), (vii) Head of ₹2 coin (2nd type), (viii) Tail of ₹2 coin (2nd type), (ix) Head of ₹5 coin (1st type), (x) Tail of ₹5 coin (1st type), (xi) Head of ₹5 coin (2nd type), (xii) Tail of ₹5 coin (2nd type), (xiii) Head of ₹10 coin, (xiv) Tail of ₹10 coin**

## 3.6 Generate Feature Vector and pass it as Input to Trained NN

In this step, a feature vector is generated from the pattern averaged coin image. The 20×20 image generates a feature vector of dimension 400×1 i.e. all the pixel values are put into a vector of 1 column. Then, this feature vector of 400 features is passed as input to trained neural network. Fig. 7 gives the architecture of Trained Neural Network.

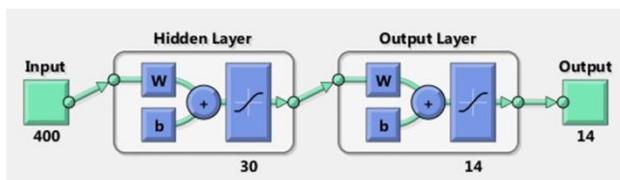

**Fig. 7: Architecture of Trained Neural Network**

## 3.7 Give Appropriate Result according to the Output of Neural Network

Coins are classified into 14 categories as shown in Fig. 2. The neural network classifies the given coin image into one of these class and based on the classification the results get generated that to which denomination the given coin belongs *i.e.* if coin gets classified in one of the class from (i) to (iv) then we say it is a ₹1 coin. Similarly, for other classes we give appropriate result. In Fig. 8 snapshot of the tool developed is given in which a ₹10 coin is recognized.

## 4. TRAINING AND TESTING DATA

Five samples of each denomination of Indian coins are scanned from both sides as shown in Fig. 2. So, it results to 10 images for each coin. But for ₹1, ₹2 and ₹5 two types of coins are used. So for each of these denominations there are 20 images from which 10 (5 for head and 5 for tail) are of 1st type and other 10 (5 for head and 5 for tail) are of 2nd type. Then after preprocessing when we get images of 100×100 then these images were rotated to $5^0$, $10^0$, $15^0$,…,$355^0$ *i.e.* total 72 rotated images get generated for each image. So there are 20*72=1440 images for each of ₹1, ₹2 and ₹5 but 10*72=720 images for ₹10. So there are total 1440*3+720=5040 images. So we trained the neural network by randomly selecting images from these 5040 images. 90% of 5040 images were used for training, and then 5% images were used for testing and rest 5% were used for validation.

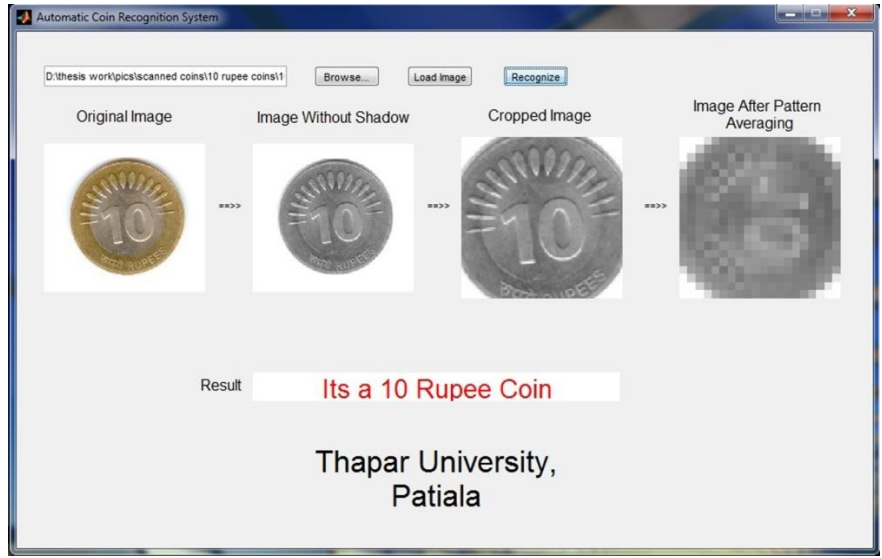

**Fig. 8: Snapshot of Automated Coin Recognition System**

## 5. RESULTS

Fig. 9 gives the resultant values of MSE (Mean Square Error) and %E obtained after training, testing and validation. Training of the network takes 148 epochs in total. Fig. 10 shows the performance of network for each training, testing and validation. The best validation performance is achieved at epoch 142. The Fig. 11 shows the confusion matrix for Neural Network. In confusion matrix Target classes are the classes to which the coin actually belongs and Output classes are the classes in which the coins get classified by trained NN. It is clear from the figure that 97.74% correct recognition has been achieved which is quite encouraging. So, there is only 2.26% misclassification.

|  | Samples | MSE | %E |
|---|---|---|---|
| Training: | 4536 | 1.65692e-3 | 2.00617e-0 |
| Validation: | 252 | 3.81927e-3 | 3.57142e-0 |
| Testing: | 252 | 6.43451e-3 | 5.55555e-0 |

**Fig. 9: Results after Training, Testing and Validation of NN**

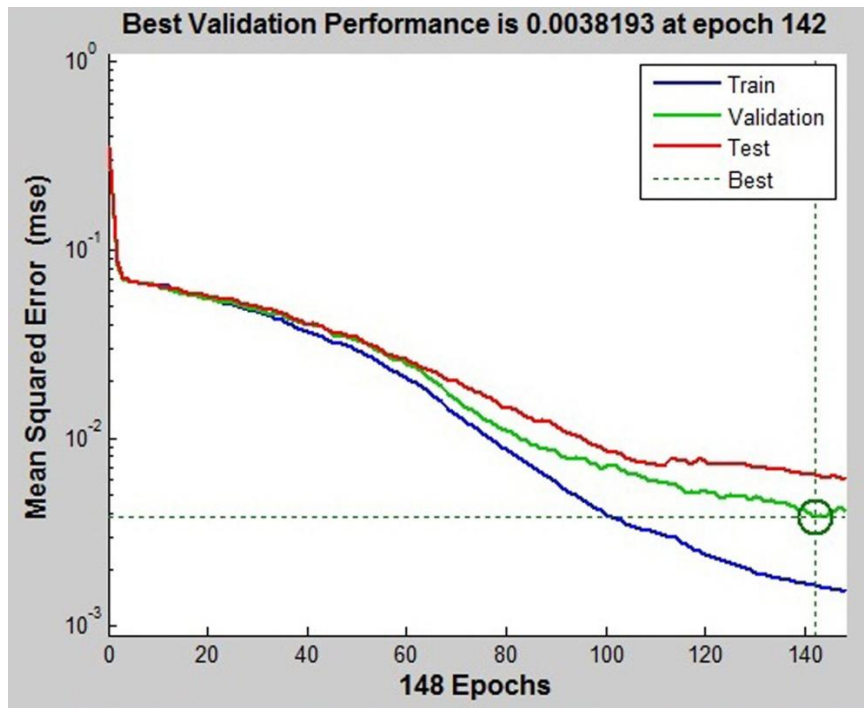

**Fig. 10: Performance of Neural Network**

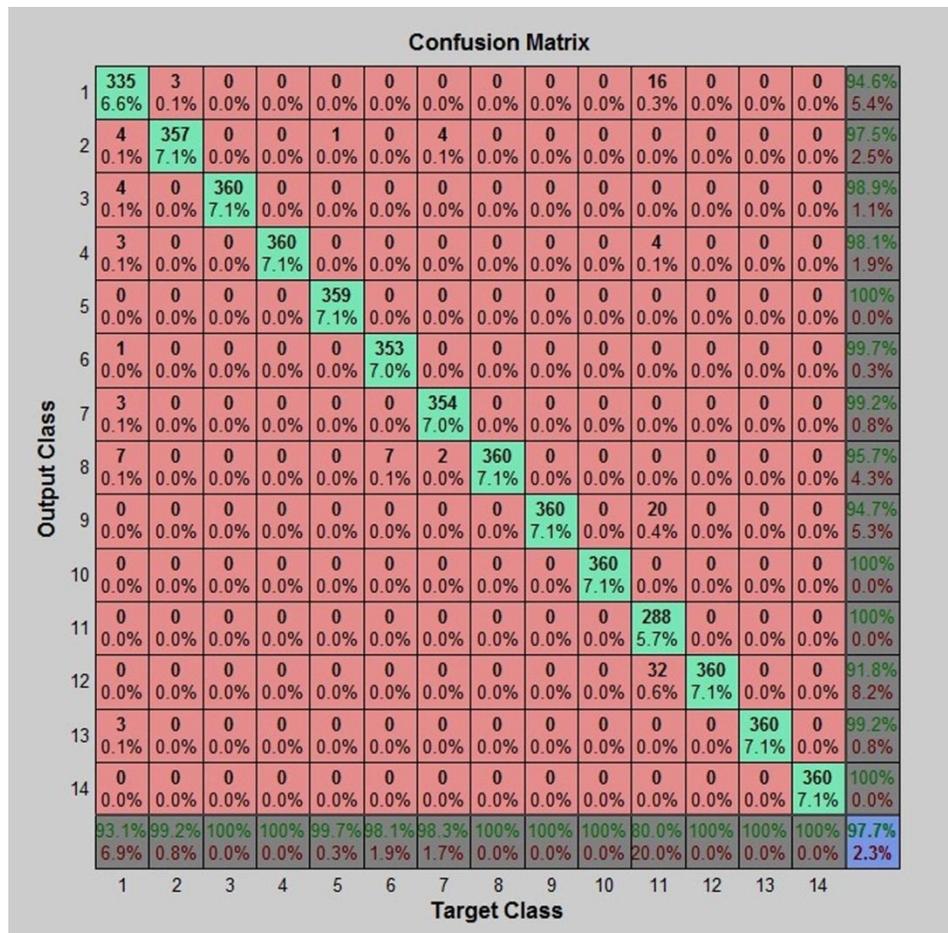

**Fig. 11: Confusion Matrix of NN**

**Table 1: Recognition Results**

| Sr. No. | Coin Type | Images correctly recognized / Total no. of images | Recognition Rate (in %age) |
|---|---|---|---|
| 1 | ₹1 | 1412/1440 | 98.05 |
| 2 | ₹2 | 1426/1440 | 99.03 |
| 3 | ₹5 | 1368/1440 | 95 |
| 4 | ₹10 | 720/720 | 100 |
| Total | | 4926/5040 | 97.74 |

## 6. CONCLUSION

An ANN based automated coin recognition system has been developed using MATLAB. In this system, firstly preprocessing of the images is done and then these preprocessed images are fed to the trained neural network. Neural network has been trained, tested and validated using 5040 sample images of denominations ₹1, ₹2, ₹5 and ₹10 rotated at $5^0$, $10^0$, $15^0$…., $355^0$. Experiments show that the system provides 97.74% correct recognition rate from 5040 sample images, *i.e.,* only 2.26% images get miss-recognized; the result is quite encouraging.

on Industrial & Engineering Applications of Artificial Intelligence & Expert Systems, 1996.

[6] Fukumi M. and Omatu S., "Rotation-Invariant Neural Pattern Recognition System with Application to Coin Recognition", IEEE Trans. Neural Networks, Vol.3, No. 2, pp. 272-279, March, 1992.

[7] Fukumi M. and Omatu S., "Designing A Neural Network For Coin Recognition By A Genetic Algorithm", Proceedings of 1993 International Joint Conference on Neural Networks, Vol. 3, pp. 2109-2112, Oct, 1993.

[8] Khashman A., Sekeroglu B. and Dimililer K., "Intelligent Coin Identification System", Proceedings of the IEEE International Symposium on Intelligent Control ( ISIC'06 ), Munich, Germany, 4-6 October 2006, pp. 1226-1230.

[9] Roushdy, M., "Detecting Coins with Different Radii based on Hough Transform in Noisy and Deformed Image", In the proceedings of GVIP Journal, Volume 7, Issue 1, April, 2007.